%% file: main_nuked.tex
\documentclass[10pt,twocolumn,letterpaper]{article}

\usepackage[pagenumbers]{iccv} 

\input{preamble}

\definecolor{iccvblue}{rgb}{0.21,0.49,0.74}
\usepackage[pagebackref,breaklinks,colorlinks,allcolors=iccvblue]{hyperref}

\def\paperID{14161} 
\def\confName{ICCV}
\def\confYear{2025}

\title{Track Anything Behind Everything: \\ Zero-Shot Amodal Video Object Segmentation}

\author{Finlay G. C. Hudson\\
Department of Computer Science \\
University of York\\
York, United Kingdom\\
{\tt\small fgch500@york.ac.uk}
\and
William A. P. Smith\\
Department of Computer Science \\
University of York\\
York, United Kingdom\\
{\tt\small william.smith@york.ac.uk}
}

\begin{document}
\maketitle
\begin{abstract}
We present Track Anything Behind Everything (TABE), a novel pipeline for zero-shot amodal video object segmentation. Unlike existing methods that require pretrained class labels, our approach uses a single query mask from the first frame where the object is visible, enabling flexible, zero-shot inference. We pose amodal segmentation as generative outpainting from modal (visible) masks using a pretrained video diffusion model. We do not need to re-train the diffusion model to accommodate additional input channels but instead use a pretrained model that we fine-tune at test-time to allow specialisation towards the tracked object. Our TABE pipeline is specifically designed to handle amodal completion, even in scenarios where objects are completely occluded. Our model and code will all be released.
\end{abstract}

\section{Introduction}

The human visual system has a strong notion of object permanence \cite{an2022object} - the idea that objects maintain their identity and continuity over time, regardless of visibility. Humans also have a notion of how objects deform or respond to other elements of an environment, allowing our minds eye to predict where an object exists and how it may have deformed, even when occluded. This concept is called amodal completion \cite{ao2023image, nanay2018importance} and is the focus of this paper.

Recent advances in computer vision have significantly improved the ability to determine an object's location within a scene based solely on its visible pixels. Methods like SAM2 \cite{ravi2024sam}, DINOv2 \cite{oquab2023dinov2}, and Mask2Former \cite{cheng2022masked} now produce segmentation masks with a level of quality and accuracy that would have seemed out of reach just a few years ago. On the other hand, a model's ability to perceive and predict the position of objects obscured by occlusions has not advanced as rapidly \cite{zhu2019robustness, kortylewski2020compositional}. Two primary challenges underlie this gap. First, obtaining real-world ground truth data for objects behind occlusions is an extremely difficult and resource intensive task \cite{vanhoorick2023tcow}. Second, defining "ground truth" itself is complex for occluded objects \cite{Zhan24}, as human perception of these hidden elements often involves educated guesses based on context, motion, or the object's anticipated behaviour. For example, if a cup obscures a ball on a table at eye level, it is ambiguous whether the ball has been fully contained by the cup or simply hidden behind it. Until the cup moves, this information remains uncertain, and even if the ball doesn't reappear, one might assume it was concealed rather than removed. This ambiguity in visual perception highlights a core problem for machine learning models. Humans can constantly update their understanding based on context and new information, while models lack this adaptive intuition. Indeed, this phenomenon is leveraged in magic tricks, where the brain's assumptions can make it susceptible to deception.

The importance of machine learning models being able to understand object permanence, particularly through amodal completion, lies in their ability to function in a manner more aligned with human cognition. Infants begin to develop a basic level of object permanence around the age of 3.5 months and continue refining this ability until about 14 months, by which point most tasks involving object permanence can be successfully solved \cite{moore1999new}. From a computational perspective, many trackers employ re-identification methodologies \cite{zheng2016mars, bergmann2019tracking, zhang2022bytetrack, zhang2021fairmot} to maintain object tracking through occlusions. However these can often be complex and non-robust. Amodal completion in video provides a human-like ability to continuously reference an object's position, even when it is behind occluders, enabling uninterrupted tracking. Furthermore, amodal completion has been shown to improve object recognition, as the ability to complete objects amodally can support accurate class recognition \cite{ozguroglu2024pix2gestalt}.

Our main contribution is as follows:
\begin{itemize}
    \item \textbf{TABE Pipeline:} We propose a pipeline for zero-shot amodal video segmentation, utilising a video diffusion model, novel target region masks and occlusion labelling technique.
\end{itemize}


\section{Related Work}

\textbf{Amodal Completion.} Amodal completion in computer vision refers to the task of inferring the parts of an object that are occluded or missing in a given image or video frame. This involves reconstructing the complete shape or structure of an object, even when parts of it are hidden by other objects or the environment. Unlike traditional segmentation, which focuses on visible parts of an object, amodal completion aims to predict the complete, unseen regions based on contextual information and the object's known properties. The approach to solving amodal completion can be categorized into two main strategies.

\textbf{Amodal Instance Segmentation} involves training a model to be able to segment both visible and non-visible portions of an object \cite{follmann2019learning, hu2019sail, mohan2022amodal}. This type of approach requires all test objects to be of classes that were seen during training time, meaning that these methods cannot be translated to a zero-shot paradigm. Within the Amodal Completion from a Visible Mask paradigm, a model must be able to amodally complete any object based on its visible mask and frame information \cite{ozguroglu2024pix2gestalt, Zhan24}. Although this method relies on extra data being provided, either as ground truth or from a modal instance segmentation model \cite{ravi2024sam, kini2024ct, oquab2023dinov2, cheng2022masked}, it allows for zero-shot amodal completion. Of particular relevance, pix2gestalt \cite{ozguroglu2024pix2gestalt} can be seen as a single image counterpart to our method. Like us, they use an outpainting diffusion model. However, their model is for single images only and they use a CLIP embedding of the whole image to provide context. In video we must handle severe or even total occlusion in some frames. We therefore found it essential to restrict the space of possible solutions using 1. our proposed target region masks and 2. fine-tuning over unoccluded observations of the object. Since fully occluded frames do not provide useful CLIP embeddings, we found they did not help.

\textbf{Video Amodal Completion.} Image-based amodal completion and segmentation struggle in the case of severe occlusion. Temporal cues play a crucial role in human ability to amodally complete objects, significantly reducing the ambiguity that often arises from partial occlusions. This skill develops early in life, with infants using spatiotemporal continuity to understand that objects persist and remain cohesive even when partially out of view. Amodal completion from visible masks can be extended for entire video sequences, as only the visible mask of the object in the first frame is required. This initial mask can be generated either through a segmentation method such as SAM2 \cite{ravi2024sam}, based on point clicks for quick object localisation, or directly from ground-truth data if available \cite{vanhoorick2023tcow}.

\textbf{Video Diffusion.} Video diffusion methods are a subset of generative models designed to generate high-quality video sequences by leveraging diffusion processes, which are originally popular in image generation tasks \cite{zhang2023survey, Rombach_2022_CVPR, ruiz2023dreambooth}. Video diffusion extends the capabilities of the image-based diffusion to handle temporal consistency across frames. Allowing for coherent videos to be generated. Video diffusion models can be conditioned in various ways, ranging from an input image to a full text description. For this task we need to ensure the video diffusion model can be conditioned by both an image and inpainting region. The CoCoCo \cite{Zi2024CoCoCo} methodology is a perfect fit for this purpose, taking both an image and inpainting region as input as well as employing global attention mechanisms to capture motion over time. This approach enables the generation of lifelike and consistent motion for query objects based on the visible mask information. Furthermore, video diffusions methods are trained on large and diverse datasets, such as WebVid-10M \cite{Bain21}, resulting in highly generalised models that are well-suited for the zero-shot paradigm we are working within. For clarity, in this paper, we refer to the process of expanding the object of interest as outpainting. However, this is achieved by inpainting a target region, using an inpainting model to generate the missing content.


\textbf{Datasets.} Existing datasets for evaluating amodal completion or segmentation are deficient in various ways. Synthetic datasets use motion tools \cite{tangemann2021unsupervised, yao2022self, gao2023coarse, girdhar2019cater, shamsian2020learning}, physics modelling \cite{vanhoorick2023tcow, HuCVPR2019, reddy2022walt, fan2023rethinking, mohan2022amodal}, or simply occlusion at an image level \cite{follmann2018mvtec, ehsani2018segan, f2007methods} and ensure perfectly accurate annotations. This allows for objects to interact, generating complex scenarios while ensuing complete control over interactions and occlusions. However, synthetic datasets suffer from the sim2real gap, where even highly sophisticated, physics-based modelling fails to fully replicate the complexities of real-world data. For example, TCOW \cite{vanhoorick2023tcow} is trained using a Kubric \cite{greff2022kubric} based dataset but performance drops significantly on real data. An alternative is to use human annotators to estimate amodal segmentation masks \cite{qi2019amodal, zhu2017semantic}. However, accuracy is limited by the annotators ability to predict missing parts and it is extremely time and labour intensive. Another alternative is to use Another alternative is to use 3D data \cite{Zhan24, li2023muva}. By capturing an object from multiple viewpoints, the full 3D shape of an object can be used to predict the full amodal mask for any view. However, 3D data acquisition is both expensive and resource-intensive and the accuracy of this approach relies heavily on the quality of the generated 3D mesh. A final option is to blend synthetic and real data via compositing, i.e. to paste an object onto a natural background to introduce occlusions. It has shown success in several other domains, including image classification \cite{walawalkar2020attentive}, image detection \cite{devries2017improved, bochkovskiy2020yolov4}, GAN's \cite{zhao2020differentiable} and visible image segmentation \cite{kini2024ct}. In amodal segmentation, pix2gestalt's \cite{ozguroglu2024pix2gestalt} training strategy employs this cut-and-paste compositing approach. However, as with similar techniques, objects in video diffusion models are often placed randomly, resulting in unrealistic occlusions that create a realism gap when applied to real-world scenarios, particularly in video data.

Due to dataset limitations, existing methods deal badly with severe or complete occlusions and this is not always reflected in the metrics. For example, TCOW \cite{vanhoorick2023tcow} provides a real video evaluation dataset in which ground truth is provided only for two unoccluded frames. Visualising the model's raw confidence heat map shows that it is highly uncertain while the object is occluded yet it performs reasonably well against the target mask metric in the final frame once the object is again visible. Such an evaluation is really evaluating re-ID rather than amodal segmentation. On the other hand, the metrics themselves do not emphasise occluded segmentation estimation. In Table \ref{tab:tcow_sam2}, we show that SAM2 \cite{ravi2024sam}, which predicts only visible, modal masks, outperforms TCOW \cite{vanhoorick2023tcow} on amodal segmentation evaluations. In other words, it is possible to score highly on amodal metrics simply by accurately tracking the visible parts of an object.

\begin{table}[h]
\centering
\begin{tabular}{lccc}
\toprule
 & \textbf{Rubric Office} & \textbf{Rubric Cup} & \textbf{Rubric DAV} \\
\midrule
TCOW & 69.4 & 38.3 & 52.8 \\
SAM2 & 72.7 & 55.8 & 68.9 \\
\bottomrule
\end{tabular}
\caption{Comparison of TCOW \cite{vanhoorick2023tcow} and SAM 2 \cite{ravi2024sam} mIoU metrics on amodal segmentation datasets. Despite SAM 2 providing only modal segmentation, it outperforms TCOW's amodal result.}
\label{tab:tcow_sam2}
\end{table}

\section{Task}

The zero-shot amodal video segmentation (ZS-AVS) task is defined as follows. We are given as input a T-frame video $x\in\mathbb{R}^{T\times H\times W\times3}$ along with a prompt of some form (e.g. text or point clicks) used to define the object of interest in some reference frame (assumed to be frame 1). The prompt is used to compute the binary query mask $m_{q}\in\{0,1\}^{H\times W}$ which segments the object of interest in the reference frame. It is assumed that the object is completely unoccluded in the reference frame. The goal is to learn a function $f$ that estimates amodal segmentation masks for the tracked object in every frame:
\begin{equation}
m_{o}=f(x,m_{q}),
\end{equation}
where $m_{o}\in\{0,1\}^{T\times H\times W}$ is the binary object mask in every frame. This must delineate the object even behind occluders and even when none of the object is visible based on information from surrounding frames and the context of the object itself.

\section{Pipeline}

We propose a novel zero-shot amodal video object segmentation method. We call this methodology TABE (Track Anything Behind Everything). The key idea is to apply outpainting to the visible regions of the tracked object using a video diffusion model.

Given an input video and point clicks to specify the target object in frame 1 (though natural language descriptions could also be used), we employ a zero-shot segmentation model to generate the query mask. We provide this query mask and the video frames to a visible VOS method (SAM2 \cite{ravi2024sam}) to compute the visible masks for each frame (which might be empty if the object is completely occluded). We use these to produce the visible masked input providing images containing only the object itself, deforming over time and with missing parts caused by occlusions. It is to these images that we apply our video diffusion outpainting method (described in Section 5), prompting the model to create a video of only the object on a white background. This provides amodal video completion output frames. Since sometimes the outpainting process reintroduces some background elements or other artefacts, we re-run VOS (SAM 2) on these frames with the original query mask, providing the final segmentation result. We found that video diffusion outpainting was liable to produce additional spurious content without restricting the region in which to outpaint (for example, if the object being tracked was a person, the outpainting result might hallucinate an additional second person). In addition, during the finetuning process for the video diffusion model, we need per-frame labels indicating which frames contain occlusion. In order to tackle both of these problems we perform occlusion reasoning prior to video diffusion. 

\textbf{Target region masks.} We restrict the diffusion outpainting to a target region mask per frame. These masks combine two cues to label likely areas the object could cover. First, we use monodepth (Depth Anything v2 \cite{depth_anything_v2}) to estimate a depth map. We compare pixel depth values against the mean depth inside the visible region mask. Any pixel with a depth value smaller than this mean is a candidate for outpainting. Second, we further restrict this candidate set of pixels by requiring that they lie within an approximate amodal bounding box. We estimate these bounding boxes using temporal continuity. We initialise the bounding boxes conservatively using the visible region masks. For frames with no visible pixels, we linearly interpolate or extrapolate from adjacent frames. If the change in bounding box area indicates a potential occlusion, we test the occlusion labels on the visible mask region boundary (see below) and grow the bounding box by assuming constant area.


\textbf{Occlusion labelling.} Intuitively, at points on the boundary of the visible mask region where the estimated depth is larger just outside the mask compared to inside, this is likely to be the true boundary of the object as opposed to an occlusion boundary. On the other hand, where the depth is smaller, it is likely that the object continues behind this closer object and so this is an occlusion boundary. Using the visible mask and estimated depth, we compute an occlusion measure which is the proportion of the boundary believed to be an occlusion boundary. Suppose that the visible mask region for a frame is defined by the set of pixels belonging to the mask $S\subset\{1,...,W\}\times\{1,...,H\}$, with boundary $\partial S$. First, we define an indicator function for a point on the boundary $(u, v)$ with outward facing normal $\vec{n}(u,v)$:
\begin{equation}
g(u,v)=\begin{cases}1&\nabla z(u,v)\cdot\vec{n}(u,v)>t\\ 0&otherwise\end{cases}
\end{equation}
 where $z(u,v)$ is the estimated depth at position $(u, v)$ and $\nabla z(u,v)\cdot\vec{n}(u,v)$ is the directional derivative of the depth function in the direction of the outward facing normal to the boundary. This quantity is positive at potential occlusion boundary points and we use $t$ slightly larger than 0 for robustness to noise. Then we compute $f_{occ}(S)\in[0,1]$ the proportion of the boundary that is potential occlusion as:
\begin{equation}
f_{occ}(S)=\frac{\int_{(u,v)\in\partial S}g(u,v)ds}{\int_{(u,v)\in\partial S}ds},
\end{equation}
 and $ds$ represents the differential arc length along the boundary to the visible mask $\partial S$. Finally, we label frame $i$ as unoccluded $(V_{i}=1)$ if $f_{occ}(S)$ is below a threshold, occluded if $f_{occ}(S)$ is above a threshold or left the frame (as indicated by bounding box extrapolation), with $V_{i}=0$ for the final two cases.

\section{Test-time Diffusion Model Fine-tuning}

Our approach relies on outpainting a partially observed object using a video diffusion model. We found that using a generic outpainting model did not lead to good results as the model did not adhere to the constraints implied on the object and motion by the partial observations. For this reason, we propose a test-time fine-tuning process in which the model is first specialised to the specific object in the input video that we wish to outpaint.

\textbf{Baseline Model.} We use the pretrained video diffusion model from COCOCO \cite{Zi2024CoCoCo}, which adapts Stable Diffusion Inpainting \cite{Rombach_2022_CVPR} for video sequences by incorporating a temporal UNet module. This temporal UNet is specifically modified to enhance motion consistency across frames, thanks to improved global information capture. The method is trained on a huge and diverse image dataset, enabling them to retain the generalisation that was possible in the original Stable Diffusion Inpainting.

\textbf{Finetuning.} We employ a similar finetuning strategy to that of Realfill \cite{tang2024realfill} but adapt it to be able to work on video data rather than just images and with occluded instances. As with Dreambooth \cite{ruiz2023dreambooth}, Realfill aims to finetune the stable diffusion model with just a few training images utilising Low Rank Adaptations (LoRA) \cite{hu2021lora} to add learnable residual modules to each weight in the network. This approach allows the primary model parameters to remain frozen, thus retaining the generalisation capability of the original diffusion model while adapting it to the new data. Further, as with Dreambooth, a fixed prompt containing a rare token is employed to specifically overfit the model to the object of interest. This allows the model to associate the token directly with the visual characteristics of that particular object.

At inference time, we need the model to output an amodal completion of the object of interest with a white background, allowing the object to be easily isolated for amodal segmentation. We begin by isolating the pixels within an estimated visible mask of the target object, setting all other pixels in the frame to white. Then, inspired by \cite{suvorov2022resolution}, we generate random binary masks that serve two purposes. First, we ensure some masks are generated that occlude parts of the object of interest. These partial occlusions encourage the model to learn how to recreate complete hidden portions of the object accurately, thus enhancing the model's amodal completion ability. Second, additional masks are randomly applied to areas outside the object of interest. This teaches the model to generate a consistent white background in those regions. These constraints ensure the model does not fill in non-object areas whilst focusing its outpainting efforts solely on reconstructing the object. Finally, while we complete full sequences of frames at each training iteration, we only optimise losses from frames which we have labelled as unoccluded via our occlusion labelling methodology. This ensures that the model receives reliable training signal for those areas without introducing noise from undefined occluded regions. Concretely, we adapt the Realfill loss as follows:
\begin{equation}
\sum_{i=1}^{T}V_{i}\cdot E_{x,t,\epsilon,p,m}(||\epsilon-\epsilon_{\theta}(x_{t},t,p,m,(1-m)\odot x)||)
\end{equation}
 where $x$ is a block of frames, $p$ is a fixed language prompt "A video of a [V] on a white background", $m$ is the randomly occluded mask, $t$ is the diffusion time step and $\epsilon$ the noise. $V_{i}$ masks the loss for occluded frames where the target outpainting is not known. Due to this being within the pipeline, as explained in Section 4, what we determine to be occluded versus unoccluded frames is not based on ground truth data apart from the initial frame where no occlusion is assumed (as defined in Section 3). Therefore, the combination of a generalised pretrained model, diverse samples of the target object, and the use of random binary masks enables the model to learn a robust representation of the object, even without perfect occlusion labels for each frame.

\section{Experiments}


\textbf{Baselines.} Existing video models typically require labelled data or training on specific object categories \cite{follmann2019learning, hsieh2023tracking}, rather than operating solely from a query image without any prior object-specific information. One exception is TCOW \cite{vanhoorick2023tcow}, a feed-forward model that, like TABE, requires only a query mask as input. The method differs slightly by also predicting occluder and container masks but we evaluate only the amodal segmentation mask. For fairness, and since we found it improved performance, we also provide the SAM2 modal segmentation masks as input. pix2gestalt \cite{ozguroglu2024pix2gestalt} performs image-based amodal completion using a diffusion model. We apply the approach independently to each video frame. SDAmodal \cite{Zhan24} relies on both visual mask and bounding box inputs. For the visual mask input, we utilise SAM2 outputs, while for the bounding box inputs, we provide ground truth amodal bounding boxes.


\begin{figure}[t]
\begin{center}
   \includegraphics[width=0.9\linewidth]{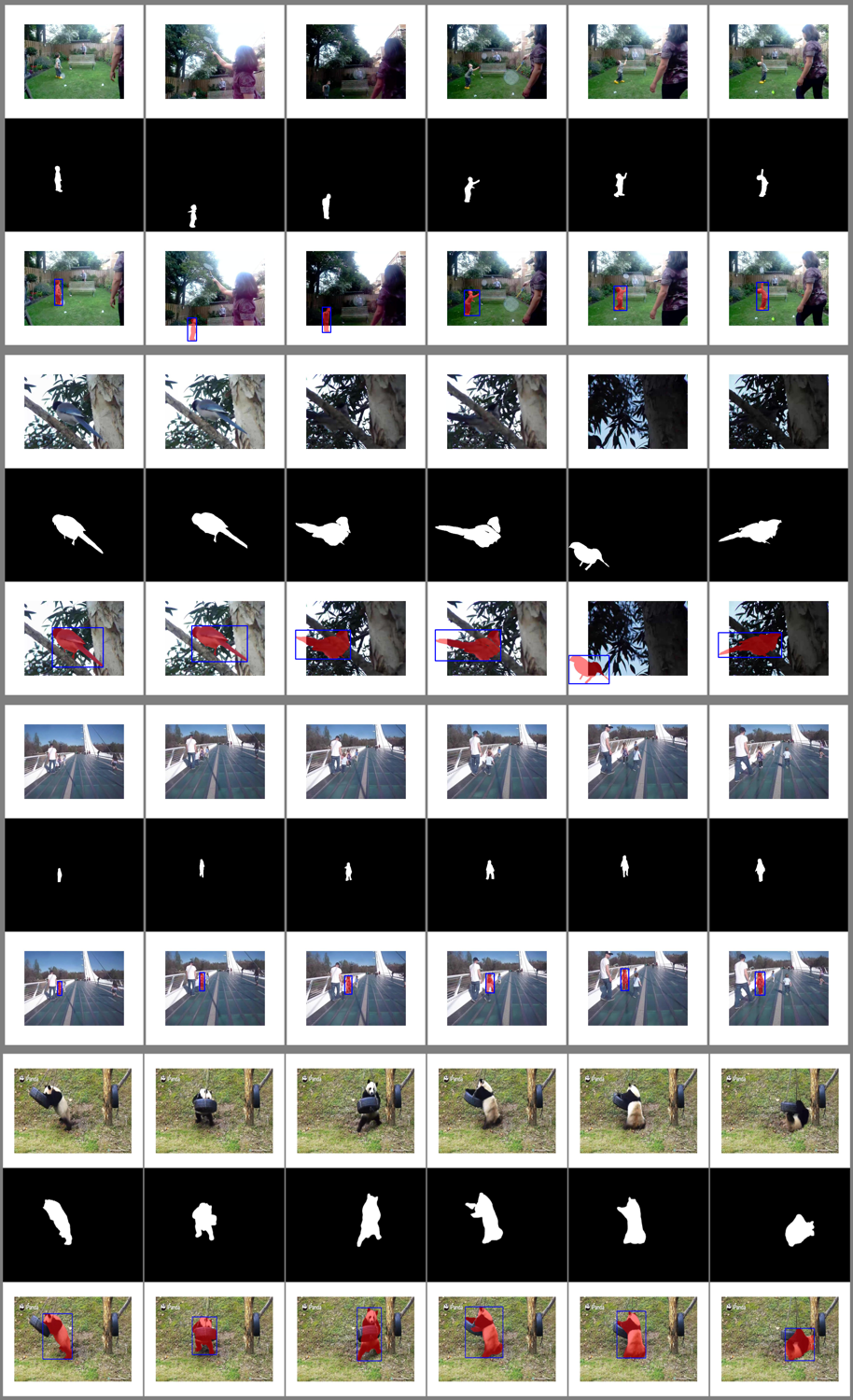}
\end{center}
   \caption{Examples of results on the TAO-Amodal dataset with our TABE method.}
\label{fig:results}
\end{figure}

\textbf{Results on TAO-Amodal.} TAO-Amodal \cite{hsieh2023tao} is a real-world video dataset of human-estimated amodal bounding boxes. To address the issue where objects are often occluded in the first frame, we created a benchmark subset consisting of 100 clips where the query object is fully visible in the first frame. The results, presented in Table \ref{tab:tao_results}, show that even though Amodal Expander \cite{hsieh2023tao} is trained on these classes, our TABE method still surpasses it by a significant margin. This highlights the effectiveness and generalization of our approach.

\begin{table}[h]
\centering
\begin{tabular}{lccc}
\toprule
\textbf{Method} & \textbf{AP@25} & \textbf{AP@50} & \textbf{AP@75} \\
\midrule
pix2gestalt \cite{ozguroglu2024pix2gestalt} & 0.223 & 0.111 & 0.043 \\
TCOW \cite{vanhoorick2023tcow} & 0.278 & 0.111 & 0.019 \\
SDAmodal \cite{Zhan24} & 0.411 & 0.330 & 0.224 \\
Amodal Expander \cite{hsieh2023tao} & 0.417 & 0.356 & 0.283 \\
TABE Estimated Bbox & 0.637 & 0.433 & 0.197 \\
\textbf{TABE (Ours)} & \textbf{0.659} & \textbf{0.518} & \textbf{0.352} \\
\bottomrule
\end{tabular}
\caption{Comparison of our TABE method across other methods on a custom TAO-Amodal benchmark subset.}
\label{tab:tao_results}
\end{table}

\section{Discussion}
Our work primarily focuses on developing a pipeline tailored for assessing amodal segmentation in videos, given only a single query frame and the video frames themselves. We describe a pipeline which provides high level results compared to other existing methods. We hope that this framework encourages further research and development within the community, building on the techniques and findings we have established.
{
    \small
    \bibliographystyle{ieeenat_fullname}
    \bibliography{refs}
}

\end{document}

%% file: preamble.tex
\definecolor{blue}{rgb}{0.0, 0.0, 1.0}
\definecolor{nicegreen}{rgb}{0.0, 0.7, 0.1}

\usepackage{booktabs}